\documentclass[conference]{IEEEtran}
\IEEEoverridecommandlockouts
\usepackage{cite}
\usepackage{amsmath,amssymb,amsfonts}
\usepackage{algorithmic}
\usepackage{graphicx}
\usepackage{textcomp}
\usepackage{xcolor}
\def\BibTeX{{\rm B\kern-.05em{\sc i\kern-.025em b}\kern-.08em
    T\kern-.1667em\lower.7ex\hbox{E}\kern-.125emX}}
    
\usepackage[T1]{fontenc}

\usepackage{caption}
\usepackage{subcaption}
\usepackage{algorithm} 
\usepackage[algo2e]{algorithm2e} 

\usepackage{url}

\makeatletter
\newcommand{\linebreakand}{%
  \end{@IEEEauthorhalign}
  \hfill\mbox{}\par
  \mbox{}\hfill\begin{@IEEEauthorhalign}
}
\makeatother
    
\begin{document}

\title{Dynamic Dispatching for Large-Scale Heterogeneous Fleet via Multi-agent Deep Reinforcement Learning
\thanks{*With equal contributions. This work has been conducted during Philip Odonkor's internship at Hitachi America Ltd.}
\thanks{$^1$ Contact: Shuai.Zheng@hal.hitachi.com}
}

\author{\IEEEauthorblockN{Chi Zhang$^*$}
\IEEEauthorblockA{\textit{Industrial AI Lab} \\
\textit{Hitachi America Ltd.}\\
Santa Clara, CA\\}
\and
\IEEEauthorblockN{Philip Odonkor$^*$}
\IEEEauthorblockA{\textit{Stevens Institute of Technology}\\
Hoboken, NJ\\}
\and
\IEEEauthorblockN{Shuai Zheng$^1$}
\IEEEauthorblockA{\textit{Industrial AI Lab} \\
\textit{Hitachi America Ltd.}\\
Santa Clara, CA\\
}
\linebreakand
\IEEEauthorblockN{Hamed Khorasgani}
\IEEEauthorblockA{\textit{Industrial AI Lab} \\
\textit{Hitachi America Ltd.}\\
Santa Clara, CA\\}
\and
\IEEEauthorblockN{Susumu Serita}
\IEEEauthorblockA{\textit{Industrial AI Lab} \\
\textit{Hitachi America Ltd.}\\
Santa Clara, CA\\}
\and
\IEEEauthorblockN{Chetan Gupta}
\IEEEauthorblockA{\textit{Industrial AI Lab} \\
\textit{Hitachi America Ltd.}\\
Santa Clara, CA\\}
}

\maketitle

\begin{abstract}
Dynamic dispatching is one of the core problems for
operation optimization in traditional industries such as
mining, as it is about how to smartly allocate the right
resources to the right place at the right time.
Conventionally, the industry relies on heuristics or even
human intuitions which are often short-sighted and
sub-optimal solutions.
Leveraging the power of AI and Internet of Things (IoT),
data-driven automation is reshaping this area.
However, facing its own challenges such as large-scale and
heterogenous trucks running in a highly dynamic
environment, it can barely adopt
methods developed in other domains (e.g., ride-sharing).
In this paper, we propose a novel Deep Reinforcement
Learning approach to solve the dynamic dispatching problem
in mining. 
We first develop an event-based mining simulator with
parameters calibrated in real mines. Then we propose an
experience-sharing Deep Q Network with a novel abstract
state/action representation to learn memories from
heterogeneous agents altogether and realizes learning in a
centralized way.
We demonstrate that the proposed methods significantly
outperform the most widely adopted approaches in the
industry by $5.56\%$ in terms of productivity. 
The proposed approach has great potential in
a broader range of industries (e.g., manufacturing,
logistics) which have a large-scale of heterogenous equipment
working in a highly dynamic environment, as a general
framework for dynamic resource allocation.
\end{abstract}


\section{INTRODUCTION}

The mining sector, an industry typified by a strong aversion to risk and change today finds itself on the cusp of an unprecedented transformation; one focused on embracing digital technologies such as artificial intelligence (AI) and the Internet of Things (IoT) to improve operational efficiency, productivity, and safety~\cite{lala2016prod}. 
While still in its nascent stage, the adoption of data-driven automation is already reshaping core mining operations. 
Advanced analytics and sensors for example, are helping lower maintenance costs and decrease downtime, while boosting output and chemical recovery \cite{mckinsey2018}. 
The potential of automation however extends far beyond. 
In this paper we demonstrate its utility towards addressing the Open-Pit Mining Operational Planning (OPMOP) problem, an NP-hard problem \cite{souza2010hybrid} which seeks to balance the trade-offs between mine productivity and operational costs. While OPMOP encapsulates a wide range of operational planning tasks, we focus on the most critical - the dynamic allocation of truck-shovel resources~\cite{chaowasakoo2017}.

\begin{figure} [t!]
\centering
           \begin{subfigure}[b]{0.35\textwidth}
         \centering
         \includegraphics[width=\textwidth]{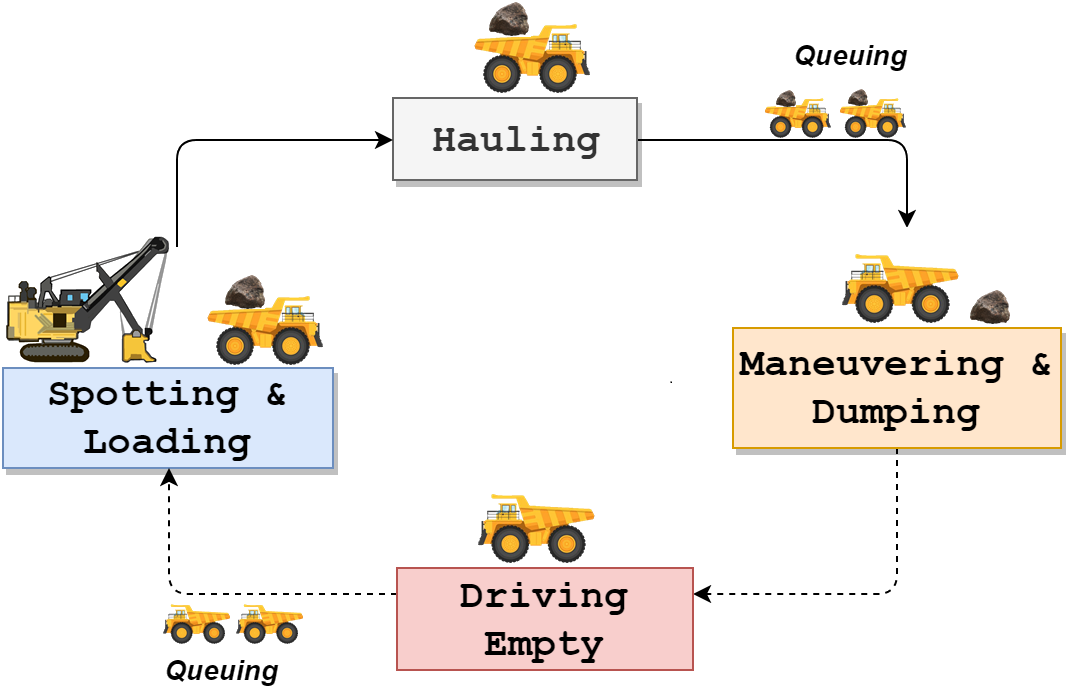} 
         \caption{}
        \label{fig:mining_cycle}
     \end{subfigure} \vspace{-5pt}
     \begin{subfigure}[b]{0.35\textwidth}
     \centering
         \includegraphics[width=\textwidth]{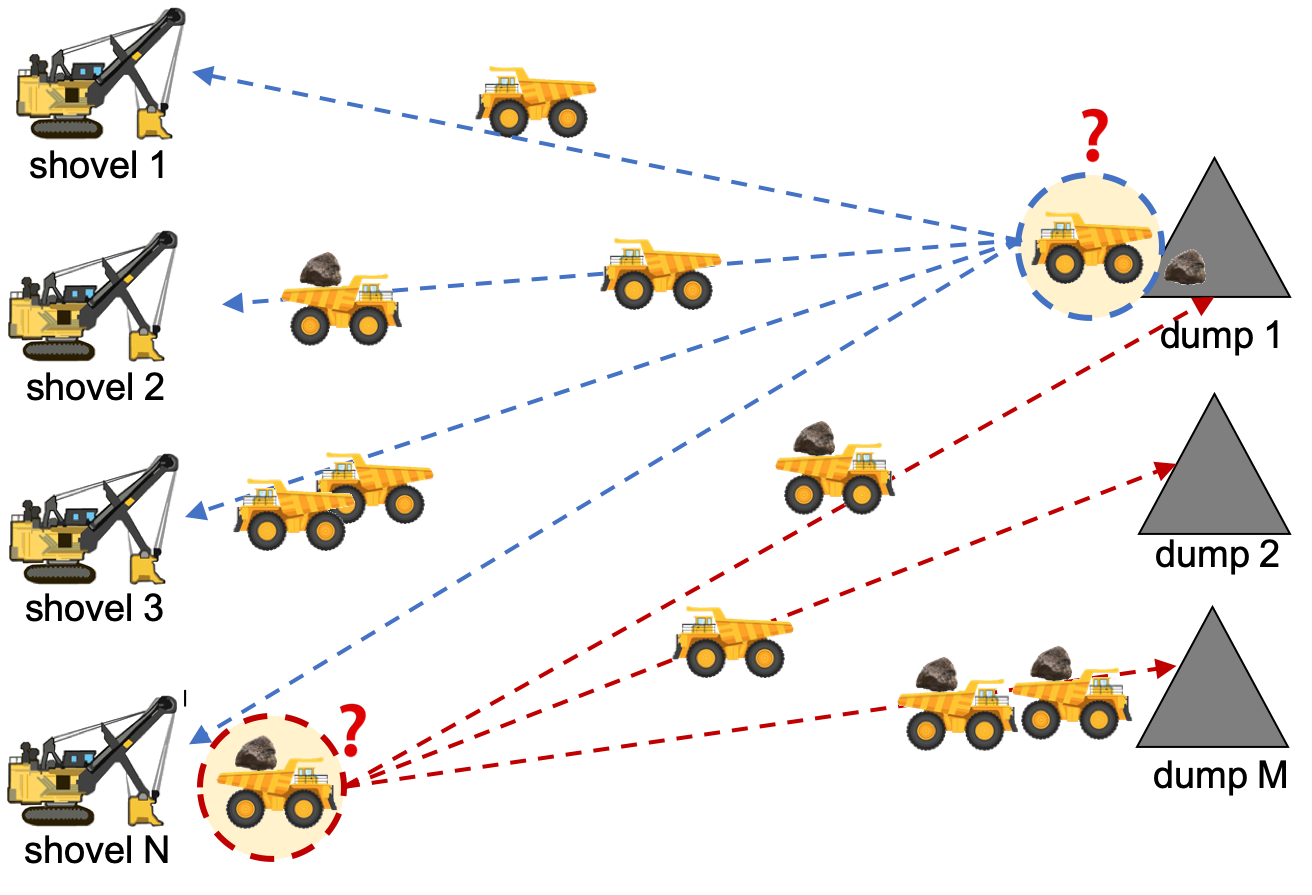}
         \caption{}
       \label{fig:dp_problem}
     \end{subfigure}
         \caption{\small{(a) Truck activities in one complete cycle in mining operations, namely driving empty to a shovel, spotting and loading, haulage, and maneuvering and dumping load. (b) Graph representation of dynamic dispatching problem in mining. When trucks finish loading or dumping (highlighted in dashed circles), they need to be dispatched to a new dump or shovel destination.}}
        \label{fig:problem}
        \vspace{-20pt}
\end{figure}

In the open-pit mine operations, dispatch decisions orchestrate trucks to shovels for ore loading, and to dumps for ore delivery. This process, referred to as a truck cycle, is repeated continually over a 12-hour operational shift. Figure~\ref{fig:mining_cycle} illustrates the sequence of events contained within a single truck cycle. An additional \textit{queuing} step is introduced when the arrival rate of trucks to a given shovel/dump exceeds its loading/dumping rate. Queuing represents a major inefficiency for trucks due to a drop in productivity. Another inefficiency worth noting occurs when the truck arrival rate falls below the shovel loading rate. This scenario is known as \textit{shovel starvation}, and result in idle shovels. Consequently, the goal of a good dispatch policy is to minimize both starvation for shovels and queuing for trucks.

With mines constantly evolving, be it through variations in fleet heterogeneity and size, or changing production requirements, open research questions still remain for developing dispatch strategies capable of continually adapting to these changes. This need is further underlined in OPMOP problems focused on dynamic truck allocation. In dynamic allocation systems, trucks are not restricted to fixed, pre-defined shovel/dump routes, instead, they can be dispatched to any shovel/dump (see Figure~\ref{fig:dp_problem}). While dynamic allocation makes it possible to actively decrease queue or starvation times, compared to fixed path dispatching strategies, this tends to be computationally more complex and demanding. In fact, efforts to address such problems using supervised learning approaches have thus far struggled to adequately capture and model the real-time changes involved~\cite{lin2018efficient}.

Particularly, there are a few challenges for dynamic dispatching in OPMOP: 1) scale of fleets are often large (e.g., a large size mine can have more than $100$ trucks running at the same time so that it is difficult for a dispatcher to make optimal decisions; 2) heterogeneous fleets with different capacities, driving times, loading/unloading speeds, etc., make it even more difficult to design a good dispatching strategy; 3) existing heuristic rules such as Shortest Queue (SQ)~\cite{subtil2011practical} and Shortest Processing Time First (SPTF)~\cite{rose2001shortest} rely on short-term and local indicators (e.g., waiting time) to make decisions, leading to short-sighted and sub-optimal solutions. In fact, the overall production performance is evaluated at shift level but the long-term and direct indicators are difficult to obtain during dispatching, if not impossible.

Recently, multi-agent deep reinforcement learning (RL) algorithms have shown superhuman performance in  game environments  such as Dota 2 video game \cite{berner2019dota} and StarCraft II \cite{vinyals2019alphastar}. In manufacturing, reinforcement learning was used for dynamic dispatching to minimize operation cost in factories\cite{zheng2019manufacturing}. In the mining industry, millions of dollars  can be saved by small  improvements in   productivity.  The unprecedented performance of multi-agent deep RL in learning sophisticated policies to win collaborative games and the huge potential benefits in the  mining industry  motivated us to investigate  the application of multi-agent deep RL  to the  dynamic dispatching challenge in this paper. 
 
In the real-world dynamic dispatching application, truck failures can happen and severely degrade the operation efficiency. On the other hand, new trucks can be put into the field at any point of time. A number of works in the area of predictive maintenance studied failure prediction or remaining useful life to reduce truck failures \cite{zheng2017long,zheng2019generative,zheng2020trace,zheng2020discriminant}. Since truck failures can happen without any warning, it is important for a robust dispatching design. Our method is robust in handling unplanned truck failures or new trucks introduced without retraining.

Section \ref{Related Work} reviews   state of the art  multi-agent deep reinforcement learning  RL algorithms. Section \ref{sec:problem} formulates the dynamic dispatching  problem as a multi-agent reinforcement learning (RL) problem. Section \ref{Experience Sharing and Memory Tailoring for Multi-agent Reinforcement Learning} presented our DQN based architecture with experience-sharing and memory-tailoring (EM-DQN) to derive optimal dispatching policies by letting our RL agents learn in a simulated environment.  In Section \ref{Mine Simulator},  we develop  a highly-configurable mining simulator with parameters learned from real-world mines to simulate trucks/shovels/dumps and their stochastic activities. Finally, we evaluate the performance of our approach against heuristic baselines that are mostly adopted in the mining industry in Section \ref{sec:experiments}. Additionally, we test out our learned models in unseen environments with truck failures to micmic the real scenarios in mines. Section \ref{Conclusions} concludes the paper.

\section{Related Work}
\label{Related Work}

When the number of agents is small, it is possible  to model multi-agents problems  using  a centralized approach \cite{silver2017mastering, mnih2013playing}  where we 
train a centralized policy  over  the agents  joint
observations  and outputting a joint set of actions. One can imagine that this approach does not scale well and quickly we will have state and action space with very large dimensions.  A more realistic approach 
 is to use an autonomous learner for each agent such as independent DQN~\cite{tampuu2017multiagent} which distinguishes agents by identities. 
Even though the independent learners  address the scalability problem   to some extend, 
they  suffer from  convergence point of view as the environments become  non-stationarity. In fact, these algorithms  model the other agents as part of the environment and, therefore,  the policy networks have to chase  moving targets as the  agents' behavior change during the training  \cite{tsitsiklis1994}.

To address the convergence problem, centralized
learning with decentralized execution approaches  have been proposed in recent years. In these methods, a  centralized learning approach is combined with a 
 decentralized execution mechanism to have the best of both worlds.  Lowe et al \cite{lowe2017multi} proposed multi-agent deep deterministic policy gradient (MADDPG), which includes   a centralized critic network and  decentralized actor networks  for the agents.  Sunehag et al \cite{sunehag2018value} proposed a linear additive  value-decomposition approach where  the total Q value is modeled as a sum of 
 individual agents' Q values.  Rashid et al \cite{rashid2018qmix} proposed  Q-MIX network which allows  a richer mixing of Q agents compared to the linear additive   value-decomposition.  Mixing network's wights are always non-negative to enforce monotonicity when it combines the agents' Q values.

Even though the centralized
learning with decentralized execution approaches have shown promising results in many applications, they are not the best candidates to address the dynamic dispatching problem for the mining industry. In the dynamic dispatching problem, the 
number of agents are not fixed. The number of available trucks can change at each given day and even when the number of trucks are known ahead of time it is fairly common for a truck to break during the operation and becomes unavailable for the rest of the  operating shift.  
Moreover, having a separate network for each truck becomes intractable from model management point of view. It is expensive to verify  models,  keep them updated, and diagnose the problems  when something goes wrong during the operation.  Unfortunately,  few mines  have access to a large data science team as a part of their operation and, therefore, simplicity and scalability  is a necessity for any application. Considering these limitation, we take the network sharing approach where    all  the agents (trucks)  share the same network, which 
 receives each agent's observation and outputs action for each agent independently. By taking this approach adding a new truck to the fleet is straightforward as the same policy has to be  applied to all the trucks.  Moreover, removing a truck does not generate a missing part in the network. Finally, the operators only have to maintain a single policy network during the operation.

  Sharing policies  among some or all of the agents has been proposed in the literature. Tan  \cite{tan1993multi} argued the agents can help each other in three main different ways; 
1) Sharing sensation, where an agent's observation   is used  by other agents to make better decisions.  2) sharing episodes, where the agents learn from each other's experiences to  
speed up learning and improve sample efficiency, and  3) sharing learned policies, where the agents take experience sharing an step further, and  share the learnt policies. Sunehag et al \cite{sunehag2018value} proposed   to force certain agents to   have the same policies by sharing weights among them in order to  avoid having  lazy (unproductive) agents in the team.

Foerster et al \cite{foerster2016learning}  proposed a  single network with shared parameters to  reduce the number of learned parameters, and speed up the learning. To address the non-stationarity problem  that
can occur when multiple agents learn concurrently, they  disabled experience replay. They argued old experiences  can become obsolete and misleading as the training carries on. Disabling the experience replay can weaken the sample efficiency and  slow down the learning process. Instead of removing experience replay altogether, we choose a more targeted approach in this paper and only remove a subset of experiments which can complicate the learning process. In the experimental results, we show that the network can converge  with the experience replay.

\section{Problem Formulation}\label{sec:problem}
Problem formulation is a very important step in applying multi-agent RL in real life applications. 
Contextual DQN (cDQN)~\cite{lin2018efficient}  reduces the number of agents by transforming the definition of agents from physical instances (i.e., all vehicles in a map) into conceptual (i.e,. coarse hexagon grids in a map) to address the online ride-sharing dispatching problem in an scalable fashion. In this paper, we consider each truck as an agent and define the same set of state variables for all the agents. This is necessary as our goal is to have the same policy network for all the agents. 

    
\subsection{Agent}

We consider any dispatchable truck as an agent. Truck fleets can be composed of trucks with varying haulage capacities, driving speeds, loading/unloading time, etc., resulting in truck fleets with heterogeneous agents.
Note that shovels and dumps are assumed to be homogeneous.

\begin{table}[b!]\vspace{-10pt}
        \centering
        \small
        \setlength\tabcolsep{2pt}\vspace{-5pt}
 \caption{Notation}
\begin{tabular}{l|l|l|l}
\hline
Pr & Meaning                                           & Pr & Meaning                 \\ \hline \hline
$T$    & Truck identity                                   	   & $s$      & A state $s \in \mathcal{S}$           \\ \hline
$C_{T}$      & Capacity of truck $T$                 & $a$      & An action $a \in \mathcal{A}$                     \\ \hline
$HL$     & Hauling time                            & $N$      & Total number of shovels \\ \hline
$DM$     & Dumping time                         & $M$      & Total number of dumps   \\ \hline
$LD$     & Loading time                           & $F$      & Total number of trucks  \\ \hline
$SP$     & Spotting time                          &     $a^{SH_n}$     & Action to go to shovel $n$                \\ \hline
$DE$     & Driving empty time                 &    $a^{DP_m}$     & Action to go to a dump $m$    \\ \hline
$TS$     & Shift duration & $M$               & Memory                  \\ \hline
 \end{tabular}
\label{table:notation}
\end{table}

\subsection{State Representation} \label{sec:state_space}

We maintain a local state $s_t$ which captures relevant attributes of the truck queues present each shovel and dump within the mine site. Particularly, when a decision (i.e., dispatching destination) needs to be made for a truck $T$, the state is represented in a vector as following:

\begin{enumerate}

    \item \textbf{Truck Capacity:} Truck capacity $C_{T}$ is captured within the state space to allow the learning agent to account for a heterogeneous truck fleet. This affords the agent the ability to develop dispatch strategies aimed at capitalizing on the capacity of trucks to maximize productivity. 
    \item \textbf{Expected Wait Time:} For each shovel and dump, we calculate the potential wait time a truck will encounter if it were dispatched to that location. To calculate this, we consider two queue types - an ``Actual Queue", $AQ$, and an ``En-route Queue", $EQ$. As the name suggests, the actual queue accounts for trucks physically queuing for a shovel or dump. The "en-route queue" on the other hand accounts for trucks that have been dispatched to a shovel or dump, but are yet to physically arrive. These two queue distinctions are necessary because they allow us to better predict the expected wait time. Consequently, the expected wait time for shovel $k$, at time $t$, $WT_t^k$, is formulated in Eqn. \ref{Eqn:waittime} as:
    \begin{equation}\scriptsize
\begin{split}
    \label{Eqn:waittime} 
        WT_{t}^{k} &= \sum_{i\in AQ^k}(LD_i + SP_i+HL_i)\\
        &+\sum_{j\in EQ^{k*}}(LD_j + SP_j+HL_j)+LD_T+SP_T+HL_T
\end{split}\vspace{-20pt}
    \end{equation}
where $LD_i$, $SP_j$ and $HL_j$ represented the average loading, spotting and hauling time of truck $i \in AQ^k$ (where $AQ^k$ is the set of all trucks in shovel $k's$ Actual Queue). The second term of this equation focuses on the En-route queue. Specifically, it is concerned with the average loading and spotting time of truck $j \in EQ^{k*}$ (where $EQ^{k*}$ is the set of all trucks in shovel $k's$ En route Queue expected to arrive \textit{before} truck $T$ if it were dispatched to this location). The following relationship always holds;  $ EQ^{k*} \leqslant EQ^{k}, \forall k$. The last two terms are the loading and spotting time of the current truck $T$. For dumps, $LD$ and $SP$ are replaced by dumping time $DM$, and $HL$ is replaced by driving empty time $DE$ in Eqn~\ref{Eqn:waittime}.
    
    \item \textbf{Total Capacity of Waiting Trucks:} For each shovel or dump we also calculate $TC_{w, t}^k$, the total capacity of all the trucks in ($AQ^k+EQ^{k*}$) which are ahead of truck $T$. This is necessary because wait time alone is not a good indicator of queue length. It is possible for a queue to have a long wait time, despite having few trucks actually queuing. Although simply providing the state space with a count of queuing trucks would have been sufficient, providing the total capacity implicitly achieved the same task, while also providing the learning agent with more useful information.   
    
    \item \textbf{Activity Time of Delayed Trucks:} Assuming our truck is dispatched to a given location, ``Delayed trucks" refer to trucks already en-route for that location which is estimated to arrive \textit{after} truck $T$. The number of delayed trucks, $DT^k$, at shovel $k$ can be derived: $ DT^k = EQ^{k} - EQ^{k*}$.
    
Based on the number of trucks in $DT^k$, the activity time, $AT$ can be calculated as follows:
    \begin{equation}\small
    \label{Eqn:DelayAct} \footnotesize
        AT_{t}^{k} = \sum_{i\in DT^k}\big(LD_i + SP_i\big)
        \vspace{-5pt}
    \end{equation}
    \item \textbf{Capacity of Delayed Trucks:} In addition to the activity time, we also calculate the combined capacity $TC_{d, t}^k$ of the delayed trucks and make this available within the state vector. Activity time and capacity of delayed trucks is included to allow the learning agent to consider the impact its decisions have on other trucks. We want the agent to be able to learn when to be selfish and prioritize its interests over other trucks, and also when to perhaps opt for a longer/slower queue for the ``greater good".
\end{enumerate}

Accordingly, 
the state of an agent $T$ at a decision making time $t$ can be represented as
\begin{equation} \label{eq:state_rep}
s_t= [C_T, <WT_t^k, TC_{w,t}^k, AT_t^k, TC_{d,t}^k>_{k=1,...,N+M}]
 \end{equation}
For a mine with $N$ shovels and $M$ dumps, the state vector length is $4 \times (N+M)+1$.
Note that when a truck needs to go to a shovel, all dumps related parts in Eqn~\ref{eq:state_rep} are masked as zeros since they have less impact on the current decision making, and vice versa for shovels. This makes the environment always ``partially observed'' by agents but effectively reduce the computational overheads.
The proposed state is different from geo-based state~\cite{lin2018efficient} or individual independent state~\cite{lauer2000algorithm, tampuu2017multiagent} with several benefits: 1) it abstracts properties among heterogeneous agents to ensure a unified representation and consequently, a centralized learning can be implemented easily (discussed in next section), 2) it is not restricted by the number of agents $F$
so that it does not need re-training when $F$ changes. This is particularly important as unplanned vehicle downtime is inevitable but re-learning is often undesired.
Note that the change of shovels and dumps are usually rare so they are assumed to be fixed.

\subsection{Action Representation}
The action space for this problem encapsulates all possible actions available to all agent. Since the dispatch problem inherently tries to determine the best shovel/dump to send a truck, each unique shovel and dump within the mine represents a possible action. Based on this approach, the action space is reduced to a finite and discrete space. The challenge of handling problems with finite, discrete action spaces is well-studied in the literature \cite{duch2007challenges,silver2017mastering}. Consequently, assuming a mine with $n$ shovels and $m$ dumps, the action space can be formulated according to Eqn. \ref{eqn.ActionSpace}. 
\begin{equation}\small
\label{eqn.ActionSpace}
 \mathcal{A} = \{a^{SH_{1}},a^{SH_{2}},...a^{SH_{N}},a^{DP_{1}},a^{DP_{2}},..., a^{DP_{M}}\}
 \vspace{-5pt}
\end{equation}

Based on this implementation, selecting an action $a^{SH_{1}}$ means that the truck in question will be dispatched to Shovel~$1$. A benefit of using this action space is that it scales very well to any number of shovels and dumps. 
It is worth noting that the only appropriate dispatch action for a truck currently at a dump is to go to a shovel, and vice versa. A truck is not allowed to go to a different dump if it is currently at a dump. The same applies to shovel locations. Consequently, part of the action space presented to an agent (see Eqn. \ref{eqn.ActionSpace}) will always be invalid.
This can however be addressed in one of two ways: (i) by awarding a large negative reward for invalid actions and ending the learning episode; or (ii) by filtering the actions. While the later approach can be easily implemented by adding simple constraints to the learner, and avoids unnecessary complexity in learning, we adopt it in this paper even though we found the first approach also works.

 \subsection{Reward Function}
The quality of each agent's action is measured via a reward signal emitted by the environment. Contrary to the norm in multi-agent RL, the reward signal is defined on an individual agent basis as opposed to being shared among agents. Since rewards are not assigned immediately following an action (owing to varying activity duration times), the approach of reward sharing becomes too cumbersome to compute. We define the individual reward $r$ associated with taking action $a$ from the reward function $\mathcal{R}(s_t, a_t)=\frac{C_{T}}{\Delta t}$, where $C_{T}$ is the capacity of truck $T$, and $\Delta t$ is the time required to complete the action $a$ (i.e., the time gap between $a_t$ and $a_{t-1}$).

\section{Experience Sharing and Memory Tailoring for Multi-agent Reinforcement Learning}
\label{Experience Sharing and Memory Tailoring for Multi-agent Reinforcement Learning}

In this section, we present a novel experience sharing multi-agent learning approach, where the learner collects state, action, and reward (i.e., experience) from each individual agent, and then learns in a centralized way.

\begin{figure}[t!]
 \includegraphics[width =0.95\columnwidth]{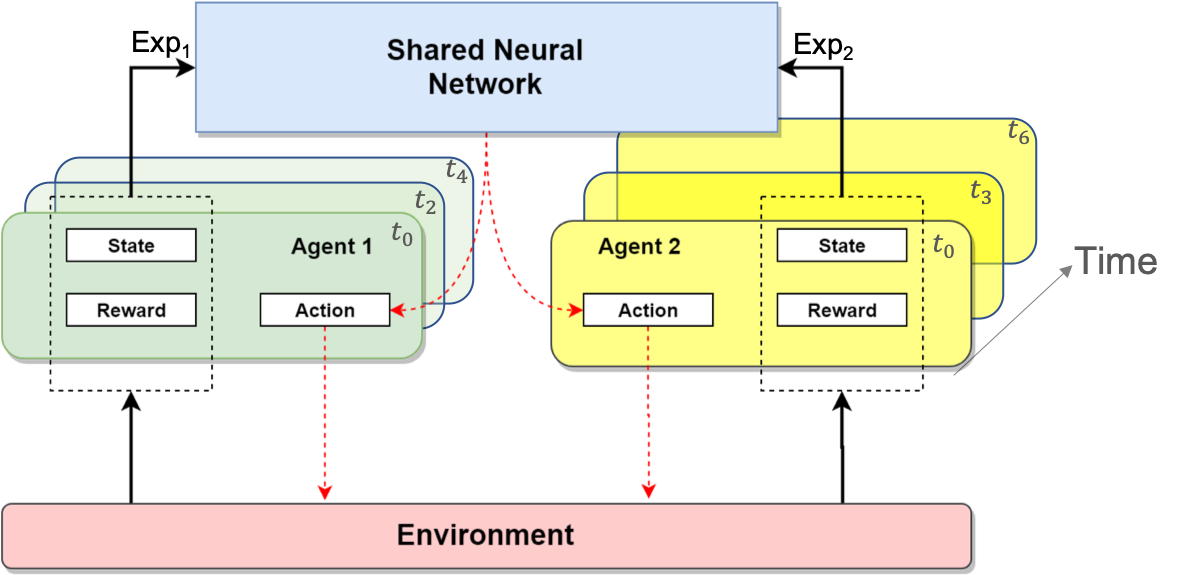} \centering
 \caption{Centralized learning with experience sharing. experience sharing by each single agent to learn an experience-sharing network.} 
 \label{fig:arch}
 \vspace{-20pt}
\end{figure}

\begin{figure}[!t]
\begin{minipage}{\columnwidth}
\begin{algorithm}[H]\small
\SetAlgoLined
\KwIn{Memory $M$; $\{T^d_{j}\}$ delayed truck IDs at shovel/dump $k$, $j=1,...DT^k$}
\KwOut{New Memory$\hat{M}$}
 Initialize Memory Tailor $MT$\;\\
 \For{$i=1$ to $k$}{
 	\For{$j=1$ to $DT^k$}{
 		$m = <S, A, S', R>_{T^d_j}$ \; \\
		$MT += m$\;
		}
 }
$\hat{M} = M - MT$\; \\
 \caption{Memory Tailoring}
 \label{alg:mem_tailoring}
\end{algorithm}
\end{minipage}
\begin{minipage}{\columnwidth}
\vspace{-5pt}
\begin{algorithm}[H] \small
\SetAlgoLined
\KwIn{state $s_t$}
\KwOut{action $a_t$}
Initialize replay Memory $M$ to capacity $M_{max}$\;\\
Initialize action value function with random weights $\theta$ \;\\
 \For{$itr=1$ to $max\textendash iterations$}{
	Reset the environment and execute simulation to obtain initial state $s_0$ \;\\
	\For{$t=0$ to $TS$}{
		\For{$i=1$ to $F$}{
			\If{Truck $T_i$ needs to be dispatched}{
				Sample action $a_t$ by $\epsilon$-greedy policy given $s_t$ \; \\
				Execute $a_t$ in simulator and obtain reward $r_t$ and next state $s_{t+\Delta{t}}$ \;\\
				Store transition $<s_t, a_t, r_t, s_{t+\Delta{t}}>_{T_{i}}$ in $M$ \;\\
				Retrieve delayed trucks ${T^d_j}$ given $s_t, a_t$\;\\
			 	Tailor $M$ using Alg.~\ref{alg:mem_tailoring} given $T^d_j$
			}
		}
	}
	\For{$e=1$ to $E$}{
		Sample a batch of transitions $<s_t, a_t, r_t, s_{t+\Delta{t}}>$ from $M$, where $t$ can be different in one batch \;\\
		Compute target $y_t=r_t+\gamma * max_{a_t+1} Q(s_{t+1}, a_{t+1}; \theta')$ \;\\
		$err = y_t - Q(s_t, a_t; \theta)$ \;\\
		Update Q-network as $\theta' \leftarrow \theta + \bigtriangledown_{\theta}e^2$
	}
}
 \caption{EM-DQN}
  \label{alg:EM-DQN}
\end{algorithm}
\vspace{-25pt}
\end{minipage}
\end{figure}

\subsection{Experience Sharing in Heterogeneous Agents}

According to Section~\ref{sec:problem}, the state and action are stored in the learner's memory without distinguishing which agent it comes from and when it is generated. 
Our key insight is that even for heterogenous agents, as long as they share the same goal and have similar functionality (i.e., all agents are trucks with loading/driving/dumping capabilities). In Section~\ref{sec:state_space}, the state space consists of truck capacity, expected wait time, total capacity of waiting trucks, activity time of delayed trucks and capacity of delayed trucks. This state representation enables abstraction of agent's properties and experience sharing becomes possible among heterogeneous agents, where the activity time such as loading, dumping and hauling (see Fig.~\ref{fig:mining_cycle}) is a function of \textit{destination type} (i.e., shovel or dump), \textit{activity type}, and \textit{fleet type}.

This makes our proposed method significantly different from previous works~\cite{lin2018efficient, lauer2000algorithm, tampuu2017multiagent} that learn multiple $Q^i$ functions where $i$ is agent identity. 

\subsection{Memory Tailoring by Coordination}

It is straightforward that every agent acts optimally based on its state at action time. Since the distance between shovels and dumps can be different, it is possible that some trucks are dispatched later than other trucks but they arrive at the shovels or dumps earlier than other trucks, if the distance is shorter. This truck will cut lines of others in this case. We identify those trucks states that are affected by this cut-line as ``corrupted" experience in the memory.
To address this problem, we propose a memory tailoring algorithm to remove the ``corrupted" experience from the memory, as shown in Alg.~\ref{alg:mem_tailoring}.

The proposed memory tailoring can be implemented by coordination mechanism, which is known to be a challenge among large-scale agents due to the high computational costs. However, in our algorithm, this overhead is small because only a small number of trucks in $EQ^{k*}$ will be affected (i.e., need to be coordinated), where $k$ is the shovel or dump ID at one time.

With the discussion above, we now present the algorithm of EM-DQN which combines experience sharing and memory tailoring in Alg.~\ref{alg:EM-DQN}.

\section{Mine Simulator}
\label{Mine Simulator}
To allow for mining dispatch operations to be simulated, a mining emulator was developed using SimPy~\cite{simpy}. SimPy is a process-based discrete-event simulation framework. Shovels and dumps were designed as resources with fixed capacity and queuing effect. 
At the point in time where a truck needs to be dispatched to either a dump or shovel, the state of all dumps and shovels are passed in as a state vector into the learner (i.e., neural network). 
The emulator enables us to quickly test different DQN architectures for developing dispatch strategies.

\begin{figure*}
\centering
  \centering
  \includegraphics[width=0.8\textwidth]{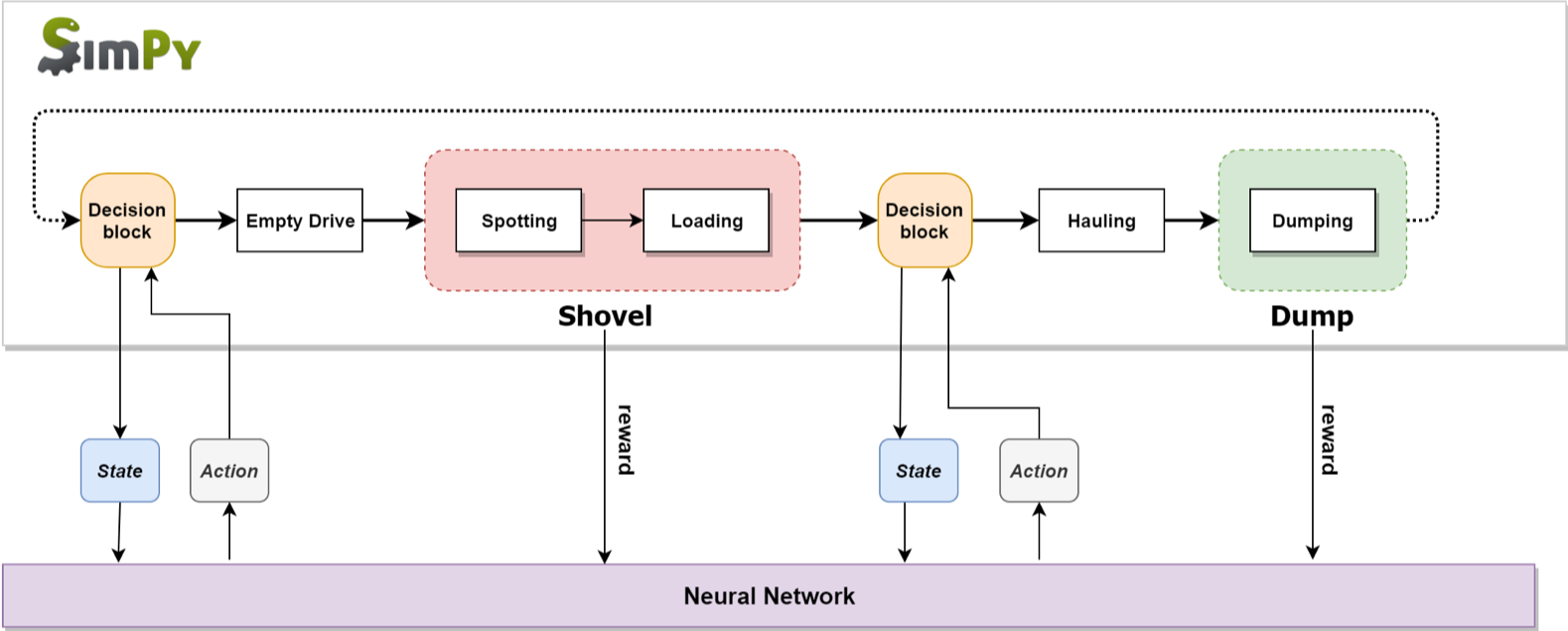}
  \captionof{figure}{Simulation framework.}
  \label{fig:simulator}
  \vspace{-20pt}
\end{figure*}

Due to that we are interested in having heterogenous fleets, the activity time such as loading, dumping and hauling (see Fig.~\ref{fig:mining_cycle}) is a function of \textit{destination type} (i.e., shovel or dump), \textit{activity type}, and \textit{fleet type}.
To increase the realism of the simulator, activity times are sampled from a set of Gamma distributions with shape and scale parameters learned from real world data in a mine we worked with~\cite{ristovski2017}. 
Fig.~\ref{fig:simulator} demonstrates the diagram of the simulator and interactions with the learner.

\section{Experimental Results}\label{sec:experiments}

To investigate the ability of the proposed method, we conduct extensive experiments to compare key metrics with heuristics that are widely adopted in mining industry.

\subsection{Experimental settings}
\subsubsection{Network settings}

The network is composed of three layers, with all followed by a ReLU\cite{nair2010rectified} activation, except the last, which has a sigmoid activation. All weights and biases are initialized according to the PyTorch default initialization.
To allow for learning, ADAM optimization algorithm is used, along with a constant learning rate of $10^{-5}$ and a batch size of 1024 samples, number of batches $E$ of $100$, memory size $M$ of $100000$, discount factor $\gamma$ of $0.9$ in Alg.~
\ref{alg:EM-DQN}. Inspired by the original DQN paper, error clipping is used to. The DQN is trained to minimize the smooth L1 loss. To encourage exploration, a simulated annealing based epsilon-greedy algorithm is used, decaying from 80\% chance of random actions down to 1\%. 

\subsubsection{Environment settings}
The training environment is set to be 3 shovels, 3 dumps, 50 trucks belonging to 3 different fleets with capacities $(200, 320, 400)$ randomly assigned to trucks. Simulation time is 12 hours.

\subsection{Baselines}
To extensively evaluate the performance of our proposed methods, we compare with baseline methods widely adopted in industries and a variance of EM-DQN:

\begin{itemize}
\item Shortest Queue (SQ) which aims to reduce cycle times is widely adopted in practice. SQ  always dispatches a truck to the destination with the minimal number of waiting trucks including those en-route trucks.

\item Smart Shortest Queue (SSQ) or Shortest Processing Time First (SPTF) take advantage of the activity time predictions to estimate the waiting time for the current truck to be served and minimize the waiting time.
We develop Smart Shortest Queue (SSQ) that makes decision to minimize the actual serving time, which is often difficult to implement for conventional SQ since it does not have the activity time estimation capability.

\item E-DQN removes the memory tailoring so that the corrupted memory is also included into model training. Using this weak version of EM-DQN, we can evaluate the effectiveness of having memory tailoring.

\end{itemize}

\begin{figure} [t!]
     \centering
     \begin{subfigure}[b]{0.4\textwidth}
         \centering
         \includegraphics[width=\textwidth]{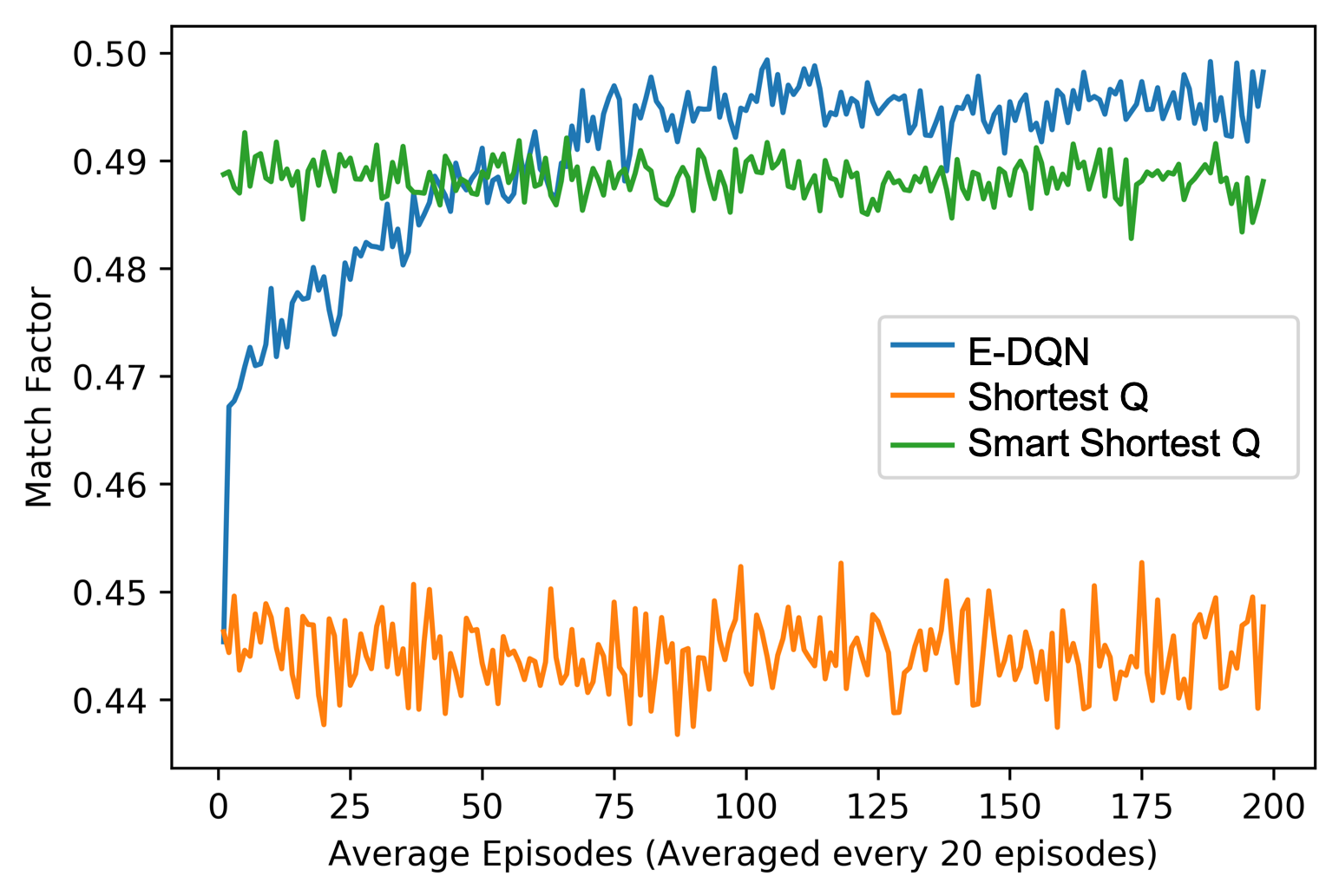}
         \label{fig:mf_2_baselines}
     \end{subfigure}
  \begin{subfigure}[b]{0.4\textwidth}
         \centering
         \includegraphics[width=\textwidth]{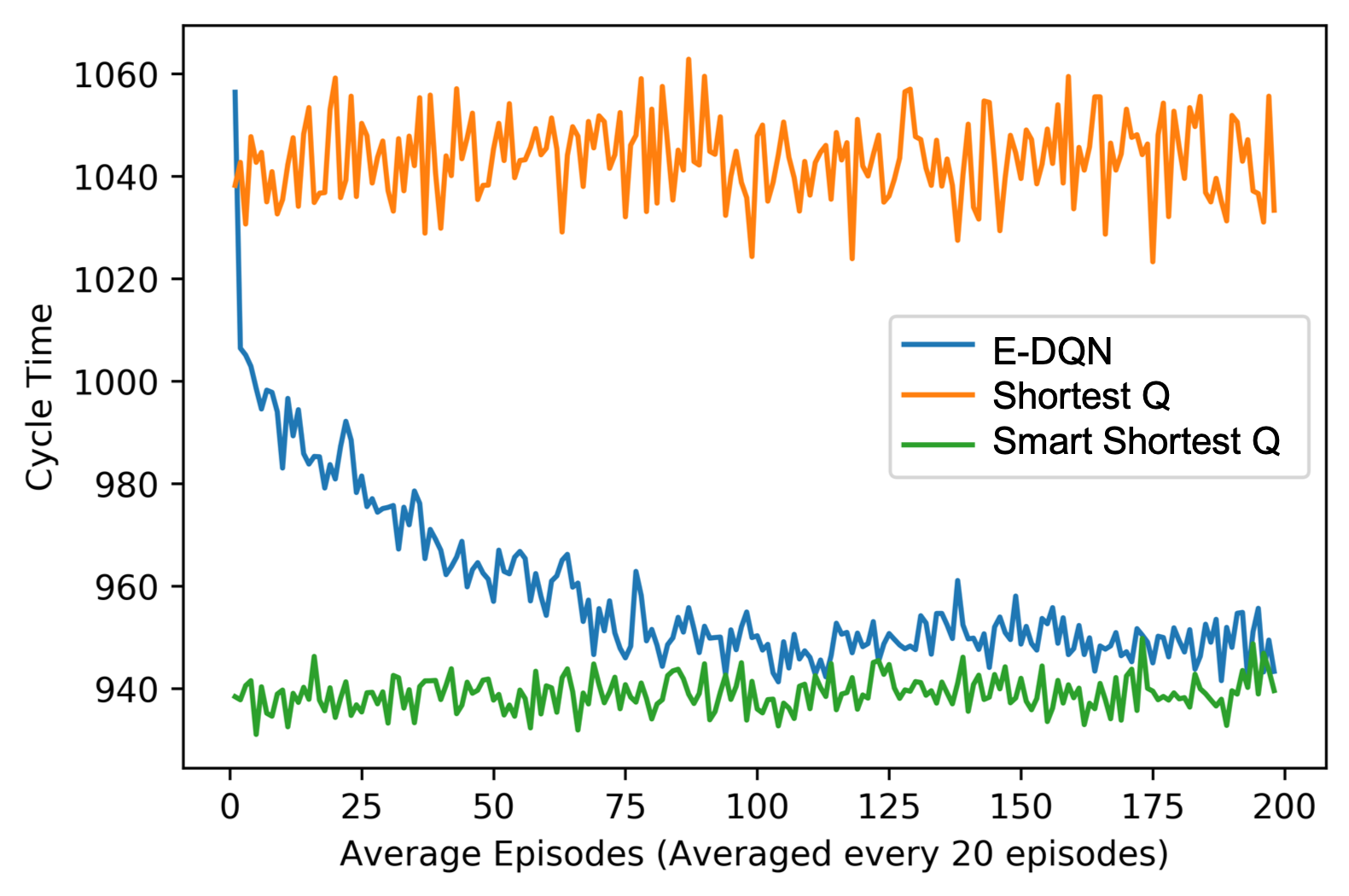}
         \label{fig:ct_2_baselines}
     \end{subfigure}
             \vspace{-10pt}
        \caption{Comparison of matching factor (left) and cycle time (right) between E-DQN, SQ, and SSQ.}
        \label{fig:compare_simple}
        \vspace{-20pt}
\end{figure}

\begin{figure} [t!]
     \centering
   \begin{subfigure}[b]{0.4\textwidth}
         \centering
         \includegraphics[width=\textwidth]{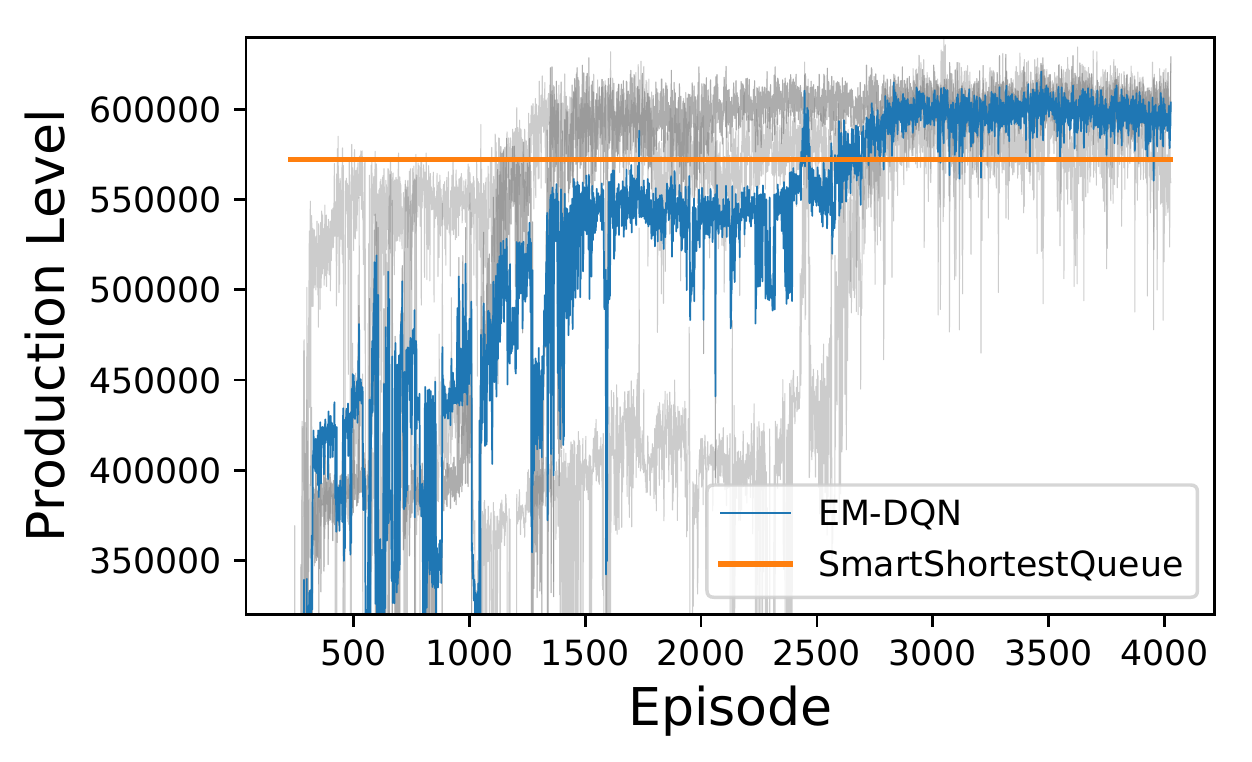}
         \label{fig:comp_prod}
     \end{subfigure}
     \begin{subfigure}[b]{0.4\textwidth}
         \centering
         \includegraphics[width=\textwidth]{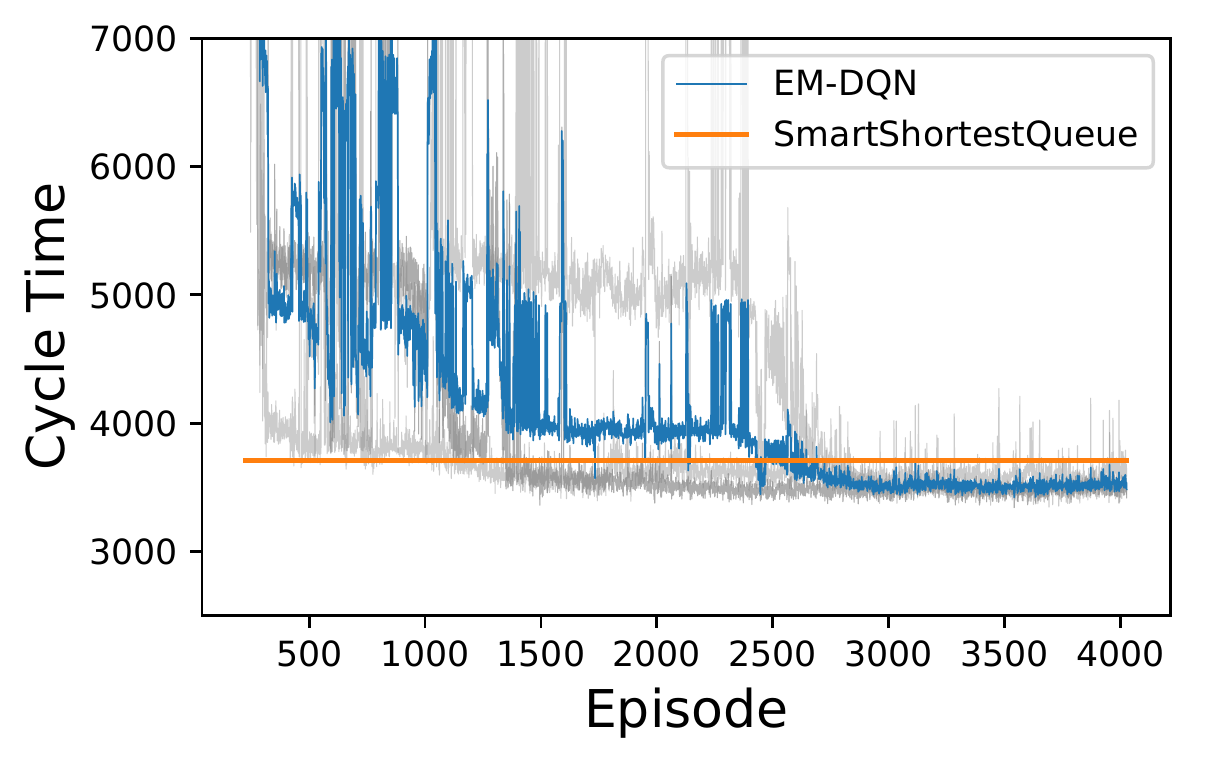}
         \label{fig:comp_ct}
     \end{subfigure}
       \begin{subfigure}[b]{0.4\textwidth}
         \centering
         \includegraphics[width=\textwidth]{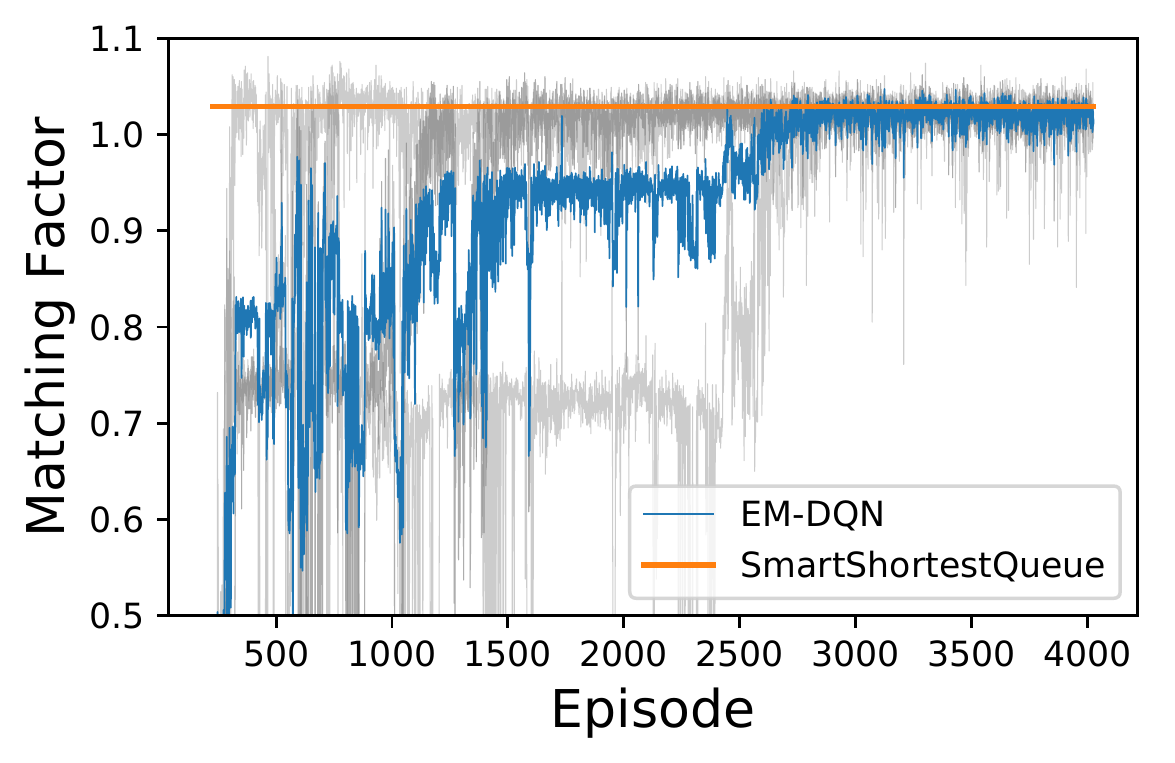}
         \label{fig:comp_mf}
     \end{subfigure}        \vspace{-20pt}
          \caption{Performance comparison of production level (left), cycle time(middle) and matching factor (right) of EM-DQN and SSQ during training. Red lines are averaged on multiple runs of EM-DQN (grey lines).}
        \label{fig:compare}
        \vspace{-15pt}
\end{figure}

\subsection{Metrics}
In the mining industry,
waiting time, idle time, utilization, queuing time, etc., are widely recognized metrics to measure the operation efficiency. 
However, these short-term metrics do not guarantee good overall performance such as production level which are long-term objectives. In this paper, we use metrics as following:
\begin{itemize}
\item \textit{Production level} is the total amount (tons) of ores delivered from shovels to dumps. This is the one of most import measurements as it is directly linked to profit mines can make. We calculate the production level in 12 hours, corresponding to one shift in mining. 
\item \textit{Cycle time} is often the short-term indicator most dispatching rules (e.g., SQ, SPTF) try to minimize. Intuitively, less cycle time yields more cycles and more delivery. However, this may not be true when we have heterogeneous trucks with different capacities. We adopt it for the purpose of comparing the short-term performance with baselines.
\item{Matching factor}~\cite{burt2007match}, which is a mid-term metric, defines the ratio of shovel productivity to truck productivity $MF=\frac{Number Of Trucks}{Number Of Shovels} \times \frac{Loading Time}{TruckCycleTime}$. Since we assume heterogeneous trucks and homogeneous shovels, the matching factor is $MF=\frac{NumberOfTrucks}{NumberOfShovels} \times \frac{\Sigma_{i}{LoadingTimeOfFleet_i \times NumberOfTrucksInFleet_i}}{\Sigma{AverageCycleTimeOfFleet_i \times NumberOfTrucksInFleet_i}}$. It is noteworthy that \small{$MF=1$} is the ideal matching of truck and shovel productivities, but it does not guarantee high production levels in heterogeneous settings. 
\end{itemize}

\subsection{Performance of EM-DQN}
We develop two types of simulated environments to evaluate the performance of our method.

\subsubsection{Cycle-based simulation}
We first use a simple environment with 3 heterogenous fleets yielding 10 trucks in total, 3 shovels, 3 dumps, and a fixed number of cycles (see Fig.~\ref{fig:mining_cycle} for the definition of one cycle) for an episode (i.e., one episode equals $k$ cycles). 
It is interesting to observe from Fig.~\ref{fig:compare_simple} that the mine is actually ``under-trucked" (i.e., $MF<1.0$), meaning that the shovel productivity is higher than truck productivity and the mine has less truck queuing but more shovel starvation. Therefore, by parallelizing the waiting time of delayed trucks and the current truck without delaying the delayed trucks, SSQ outperforms SQ significantly in terms of cycle time and E-DQN has a close performance as SSQ, as shown in Fig.~\ref{fig:compare_simple}. However, E-DQN outperforms SSQ in terms of matching factors as it achieves more balanced $MF$.

\begin{figure*}[h]  \vspace{-20pt}
\centering
  \centering
  \includegraphics[width=0.6\textwidth]{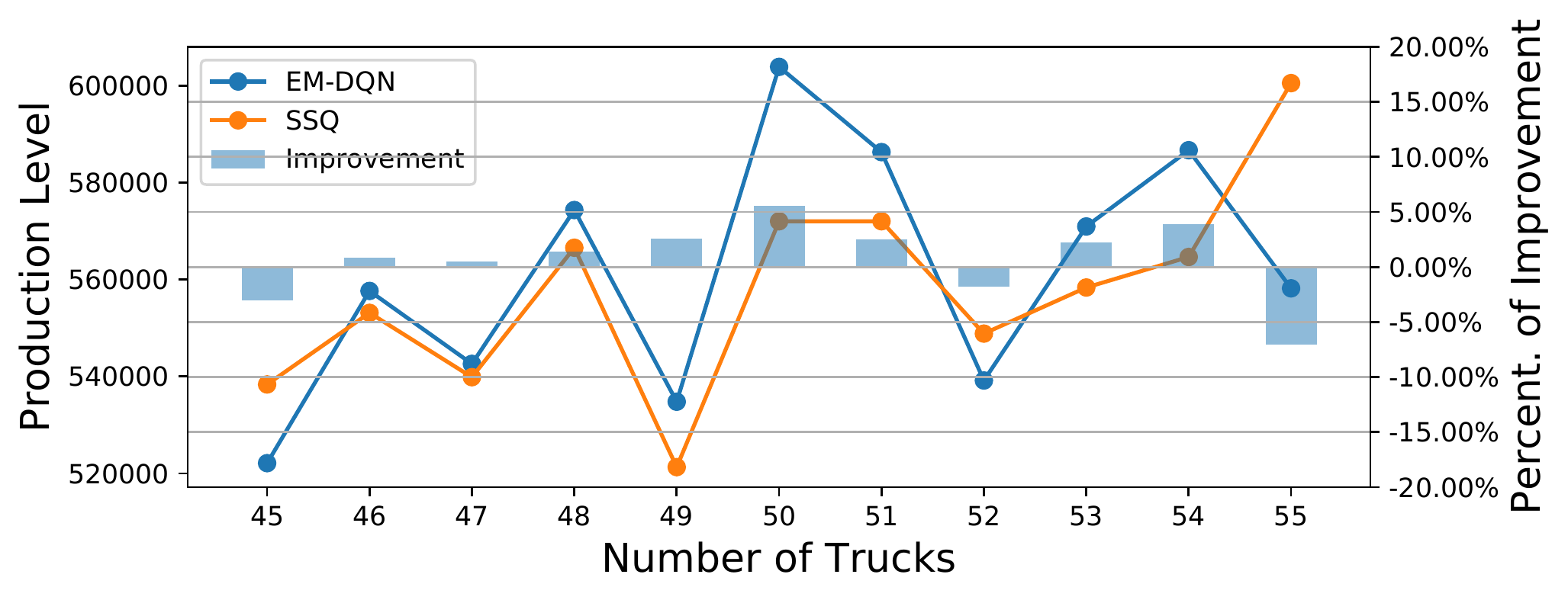}
          \vspace{-10pt}
  \captionof{figure}{\small{Testings in new environments with various number of trucks.}}
  \label{fig:test}
\end{figure*}

\subsubsection{Time-based simulation}\label{sec:time_based}
Scaling from the small problem settings (i.e., 10 trucks and 10 fixed cycles only), a more complex environment is created as an approximation of a real mine we worked with before. It has 50 trucks belonging to 3 heterogeneous fleets, 3 shovels, 3 dumps, and is simulated in based on time (i.e., 12 simulation hours corresponding to one shift). Since we already know SSQ is much better than SQ due to the activity time estimation capability, in this experiment we report performance comparison between EM-DQN and SSQ only. It is observed that now the mine is ``over-trucked'' ($MF>1$ in Fig.~\ref{fig:compare}), while EM-DQN is slightly better (i.e., balanced) than SSQ. A plausible explanation is that as the number of trucks increases from 10 to 50, queuing becomes a severe problem. In this environment, we train EM-DQN multiple times and show the training process in Fig.~\ref{fig:compare}. It can be observed that after around 10000 episodes, EM-DQN outperforms SSQ in terms of all three metrics.
On average, EM-DQN produces $603840.0$ tons compared with SSQ at $572016.87$ tons. Therefore, $31823.13$ tons of more ore can be delivered during a shift, which is $\frac{31823.13}{572016.87} \approx \bold{5.56}\%$ improvement. In fact, it shows by just dispatching trucks with EM-DQN models, $\bold{79.55}$ more free cycles can be achieved for the trucks with the maximum capacity of 400 tons.

\subsection{Robustness}
Our proposed RL approach is robust to truck failures in this design. To mimic such unexpected situations and validate the robustness of our method, we test out the model learned in Section~\ref{sec:time_based} with a series of new environments with various number of trucks (i.e., $45\sim 55$), where the failed or added trucks are randomly selected from the heterogenous fleets.

Fig.~\ref{fig:test} shows that even though the model is trained in the $50$ trucks environment, it can still maintain high and stable production levels for environments with a wide range of different number of agents  (i.e., $\pm10\%$).
Note that this is achieved \textit{without} re-training the model, which distinguishes our method from previous works~\cite{tampuu2017multiagent, lin2018efficient} where re-training is needed when the number of agents changes.
Additionally, it can also be observed that EM-DQN outperforms SSQ in 8 out of 10 testing environments. 
As a result, it demonstrates that EM-DQN can generate highly efficient dynamic dispatching policies with good robustness.


\section{Conclusions}
\label{Conclusions}

Dynamic dispatching is crucial in industrial operation optimization. Due to the complexity of the mining operations, it remains a difficult problem and still relays heavily on rule-based approaches.
This paper takes a
major step forward toward by formulating this problem as a MARL problem.
We first develop a highly-configurable event-based mining simulator with parameters learned from real mines we worked with before. 
Then we propose EM-DQN method to realize an efficient centralized learning. We demonstrated the effectiveness of the proposed method on by comparing it with the most widely adopted baselines in mining industry. 
We showed that our method can significantly improve the production level by $5.56\%$, which equals to $79.55$ more free cycles of the largest capacity truck per shift.
Particularly, our method is robust in handling unplanned truck failures or new trucks introduced without retraining. We believe this makes our method more useful for the industry as such unexpected events happen very often in real mines. 


\bibliographystyle{ieeetr}
\bibliography{bibliography_v2}

\end{document}